\documentclass[11pt]{article}

\usepackage[margin=1in]{geometry}
\usepackage{times}
\usepackage{microtype}
\usepackage[numbers,sort&compress]{natbib}
\usepackage{hyperref}
\usepackage{url}
\usepackage{booktabs}
\usepackage{tabularx}
\usepackage{enumitem}
\usepackage{xcolor}
\usepackage{listings}
\usepackage{graphicx}

\hypersetup{
  colorlinks=true,
  linkcolor=blue,
  citecolor=blue,
  urlcolor=blue
}

\title{
\begin{tabular}{c@{\hspace{0.8em}}c}
\raisebox{-0.35\height}{\includegraphics[width=0.08\textwidth]{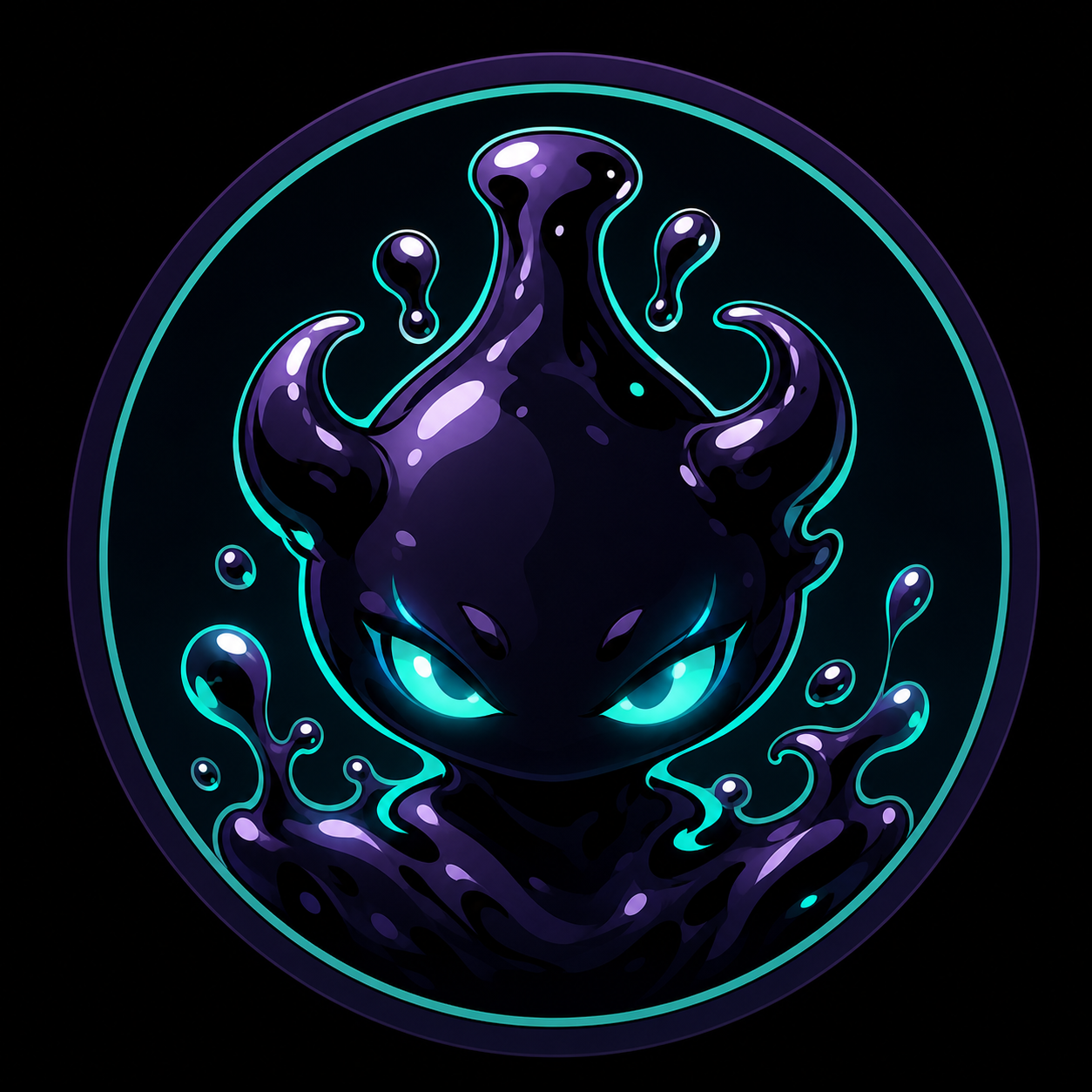}}
&
\textbf{Venom: A PyTorch Generative Modeling Toolkit}
\end{tabular}
}

\author{
Liang Yan\\
Paul G. Allen School of Computer Science \& Engineering \\
University of Washington\\
\texttt{https://divinyan.com/Venom/} \\
\texttt{https://github.com/yanliang3612/Venom}
}

\date{}

\begin{document}
\maketitle

\begin{abstract}
Modern generative modeling has grown into a broad collection of related but
often separately implemented paradigms, including denoising diffusion models,
score-based stochastic differential equations, flow matching, variational
autoencoders, normalizing flows, adversarial models, and energy-based models.
For newcomers, this fragmentation makes it difficult to compare training
objectives, inference procedures, sampling algorithms, and conditioning
mechanisms within a single coherent codebase. We introduce \textsc{Venom}, an
educational PyTorch toolkit that implements representative generative modeling
families under a unified, MNIST-first interface. \textsc{Venom} emphasizes
breadth, readability, reproducible entry points, and consistent training and
sampling APIs rather than large-scale performance engineering. The package
currently includes diffusion and score-based models, flow matching and
one-step generators, variational autoencoders, normalizing flows, generative
adversarial networks, and energy-based models. It provides separate training
and sampling scripts, classifier and classifier-free guidance examples,
bilingual tutorial notebooks, and a model-family organization that supports
teaching, prototyping, and lightweight benchmarking.
\end{abstract}

\section{Introduction}

Generative modeling has become a central area of modern machine learning.
Diffusion models \citep{ho2020ddpm,nichol2021improved,dhariwal2021diffusion},
score-based models \citep{song2019ncsn,song2021score}, variational autoencoders
\citep{kingma2014vae}, normalizing flows
\citep{rezende2015normalizing,dinh2017realnvp}, generative adversarial networks
\citep{goodfellow2014gan}, and energy-based models
\citep{lecun2006energy,du2019implicit} provide different ways to learn,
sample from, and condition on complex data distributions. Although these
families are often studied separately, they address closely related questions:
how should probability distributions be represented, how should samples be
generated, and how should conditioning or guidance be incorporated?

This diversity has greatly advanced the field, but it also creates a practical
barrier for learning, comparison, and prototyping. Research code is often
designed around a specific paper or model family, while tutorials typically
focus on one paradigm at a time. As a result, newcomers may learn the mechanics
of diffusion training without seeing how it relates to adversarial learning,
exact-likelihood normalizing flows, variational inference, or energy-based
sampling. Similarly, researchers and instructors who want a compact benchmark
or teaching scaffold often need to combine multiple repositories with different
APIs, dataset conventions, training scripts, and checkpoint formats.

\textsc{Venom} is designed to address this gap. It collects representative
algorithms from major generative modeling families into a single lightweight
PyTorch codebase with a unified, MNIST-first interface. Rather than aiming to
reproduce every engineering detail of large-scale image generation systems,
\textsc{Venom} emphasizes breadth, readability, and consistency. Its goal is to
provide a compact and inspectable map of common training objectives, inference
procedures, sampling algorithms, and guidance mechanisms. By using MNIST as a
shared starting point, each model family can be trained quickly, inspected
easily, and extended to more complex datasets or research settings.

\section{Design Goals}

\textsc{Venom} is designed around four principles: clear organization,
consistent entry points, educational coverage, and lightweight benchmark
readiness.

\paragraph{Family-level organization.}
Each major generative modeling paradigm is implemented in a dedicated
subpackage, including \texttt{venom.diffusion}, \texttt{venom.vae},
\texttt{venom.flows}, \texttt{venom.gan}, and \texttt{venom.ebm}. This
structure makes the connection between the codebase and the underlying modeling
families explicit, allowing users to navigate the repository by conceptual
paradigm rather than by implementation details.

\paragraph{Consistent training and sampling entry points.}
We standardize the experimental interface across model families by providing
paired training and sampling scripts for each method. For example, diffusion
models expose \texttt{train\_diffusion.py} and
\texttt{sample\_diffusion.py}, while normalizing flows expose
\texttt{train\_flow.py} and \texttt{sample\_flow.py}. This consistent structure
reduces the overhead of reproducing experiments and makes it easier to compare
different model families under the same workflow. Users can therefore run
meaningful experiments through simple command-line entry points without first
learning a large or highly specialized framework.

\paragraph{Educational completeness.}
The package intentionally covers a broad range of training and inference
paradigms, including variational lower bounds, adversarial objectives, exact
likelihood maximization, score matching, contrastive divergence, Langevin
sampling, ODE/SDE-based sampling, and classifier guidance. The goal is not only
to provide working implementations, but also to help users understand how these
paradigms relate to one another within a shared experimental setting.

\paragraph{Benchmark readiness.}
Because the implemented families share a default dataset, checkpoint convention,
and sampling interface, \textsc{Venom} can also serve as a lightweight benchmark
for comparing objectives, samplers, and conditioning mechanisms. This makes the
package useful not only for learning, but also for controlled prototyping and
small-scale empirical comparison.

\section{Implemented Model Families}

Table~\ref{tab:coverage} summarizes the current model coverage in
\textsc{Venom}. The package does not aim to provide state-of-the-art
reproductions of every original method. Instead, it offers compact and
inspectable implementations of the core ideas behind each model family, making
them suitable for learning, prototyping, and controlled benchmarking.

\begin{table}[t]
\centering
\small
\caption{Model families covered by \textsc{Venom}.}
\label{tab:coverage}
\begin{tabularx}{\textwidth}{p{0.22\textwidth} p{0.28\textwidth} X}
\toprule
Family & Implemented variants & Core idea represented \\
\midrule
Diffusion and score-based models &
DDPM, Improved DDPM, ADM, CFG, DDIM, DPM-Solver, Score SDE, NCSN, EDM, PFGM &
Denoising objectives, score estimation, SDE/ODE sampling, and classifier or classifier-free guidance
\citep{ho2020ddpm,nichol2021improved,dhariwal2021diffusion,song2020ddim,
song2019ncsn,song2021score,karras2022edm,lu2022dpmsolver,xu2022pfgm}. \\
\addlinespace
Flow matching and one-step models &
Rectified Flow, Flow Matching, Conditional Flow Matching, stochastic interpolants, consistency models, shortcut models, MeanFlow &
Velocity-field learning, trajectory-based generation, and fast few-step or one-step generation, including consistency-style distillation
\citep{liu2023rectifiedflow,lipman2023flow,tong2024conditionalflowmatching,
albergo2023stochastic,song2023consistency,salimans2022progressive,
frans2025shortcut,geng2025meanflow}. \\
\addlinespace
Variational autoencoders &
VAE, ConvVAE, $\beta$-VAE, CVAE, IWAE, VQ-VAE, Ladder VAE, Flow-VAE &
Latent-variable modeling, amortized inference, importance weighting, discrete latent representations, and flow-based posterior corrections
\citep{kingma2014vae,burda2016iwae,higgins2017beta,oord2017vqvae,rezende2015normalizing}. \\
\addlinespace
Normalizing flows &
Planar, radial, NICE, RealNVP, Glow, MAF, IAF, Neural Spline Flow, FFJORD, Flow++ &
Invertible transformations, change-of-variables likelihoods, and exact density estimation
\citep{rezende2015normalizing,dinh2014nice,dinh2017realnvp,kingma2016iaf,
kingma2018glow,papamakarios2017maf,durkan2019spline,grathwohl2019ffjord,
ho2019flowpp}. \\
\addlinespace
Generative adversarial networks &
GAN, DCGAN, CGAN, ACGAN, InfoGAN, LSGAN, WGAN, WGAN-GP, HingeGAN, SNGAN &
Adversarial learning, conditional generation, Wasserstein objectives, gradient penalties, hinge-loss training, and spectral normalization
\citep{goodfellow2014gan,radford2016dcgan,mirza2014cgan,odena2017acgan,
chen2016infogan,mao2017lsgan,arjovsky2017wgan,gulrajani2017wgangp,
lim2017geometricgan,miyato2018sngan}. \\
\addlinespace
Energy-based models &
RBM, Gaussian RBM, Conditional RBM, ConvRBM, Deep EBM, Conditional EBM, JEM, score matching EBM, sliced score matching EBM, NCE-EBM &
Explicit energy functions, contrastive divergence, Langevin/SGLD sampling, score matching, and noise-contrastive estimation
\citep{hinton2002cd,hyvarinen2005score,gutmann2010nce,du2019implicit,grathwohl2020jem}. \\
\bottomrule
\end{tabularx}
\end{table}

\section{Package Structure and API}

\textsc{Venom} provides both command-line entry points and Python APIs. The
command-line interface is designed for quick experiments and reproducible
examples, while the Python API supports extension, modification, and research
reuse.

\begin{lstlisting}[language=bash,caption={Example command-line workflow.}]
pip install -e .
python train_diffusion.py --variant ddpm --epochs 5
python sample_diffusion.py --checkpoint runs/mnist_diffusion/ddpm/model_005.pt

python train_vae.py --variant conv-vae --epochs 5
python train_flow.py --variant realnvp --epochs 5
python train_gan.py --variant dcgan --epochs 5
python train_ebm.py --variant rbm --epochs 5
\end{lstlisting}

The Python API follows a family-specific factory pattern:

\begin{lstlisting}[language=Python,caption={Example Python API usage.}]
from venom.diffusion.factory import build_mnist_diffusion
from venom.flows import build_mnist_flow
from venom.gan import build_mnist_gan

model, diffusion = build_mnist_diffusion("ddpm")
flow, flow_config = build_mnist_flow("realnvp")
generator, discriminator, gan_config = build_mnist_gan("dcgan")
\end{lstlisting}

This design keeps the modeling assumptions visible rather than hiding them
behind a single generic interface. For example, normalizing flows expose
likelihood-oriented methods such as \texttt{log\_prob}; diffusion models expose
different sampler choices such as DDPM, DDIM, and DPM-Solver; and EBMs expose
Gibbs, Langevin, or SGLD-style sampling procedures. As a result, users can
interact with each family through an interface that reflects its underlying
mathematical structure while still benefiting from a consistent repository-level
workflow.

\section{Training, Inference, and Guidance Paradigms}

A central goal of \textsc{Venom} is to place different generative modeling
paradigms within the same codebase and make their operational differences
explicit.

\paragraph{Training.}
Different model families rely on different training principles. Diffusion
models train denoisers or score networks; VAEs optimize variational bounds;
normalizing flows maximize exact likelihoods under invertible transformations;
GANs optimize adversarial objectives; and EBMs learn scalar energy functions
using contrastive, score-based, or noise-contrastive estimators. These
objectives are often introduced in separate literatures, but in \textsc{Venom}
they can be studied with comparable data loaders, training loops, checkpoint
structures, and sampling outputs.

\paragraph{Inference and sampling.}
Sampling also differs substantially across model families. Diffusion models
reverse a stochastic or deterministic denoising process; score SDEs use
predictor-corrector or ODE-based samplers; normalizing flows transform Gaussian
latent variables through an invertible map; GANs generate samples through a
single forward pass; VAEs decode latent variables; and EBMs use MCMC-style
Gibbs, Langevin, or SGLD transitions. \textsc{Venom} exposes these differences
through separate sampling scripts and family-specific APIs, making the
computational behavior of each paradigm easier to inspect and compare.

\paragraph{Guidance and conditioning.}
The package includes class conditioning, classifier guidance, and
classifier-free guidance for diffusion-style models
\citep{dhariwal2021diffusion,ho2021classifierfree}. It also includes
label-conditioned variants of VAEs, GANs, and EBMs. This allows users to compare
modern guidance mechanisms with simpler conditional generation paradigms in a
shared experimental environment.

\section{Tutorials and Benchmark Use}

The repository includes both Chinese and English Jupyter notebooks that walk
through training, sampling, guidance, checkpoint locations, and smoke-test
performance indicators. These tutorials are intended to make the package
accessible to entry-level users, students, and researchers who want a compact
overview of modern generative modeling methods.

The same scripts can also be used as a small benchmark suite. Each model family
writes checkpoints and sample grids under a separate \texttt{runs/} directory,
which makes it straightforward to compare training behavior, sampling behavior,
and qualitative outputs across different objectives and samplers. Although
\textsc{Venom} is not designed as a large-scale benchmark, its unified interface
makes it useful for lightweight empirical studies and teaching-oriented
comparisons.

\section{Scope and Limitations}

\textsc{Venom} is not intended to replace specialized high-performance
implementations. Large-scale diffusion systems, high-resolution flow models,
and industrial GAN training pipelines require additional engineering, data
infrastructure, distributed training, and evaluation tools. Instead,
\textsc{Venom} targets the gap between isolated tutorials and large research
repositories. Its emphasis is on coverage of core paradigms, conceptual clarity,
readability, and ease of modification.

The current implementation is centered on MNIST to keep experiments fast and
accessible. While this choice makes the package suitable for education,
debugging, and controlled comparison, it also means that the default setting is
not intended to measure large-scale generative modeling performance. Future
extensions could include larger datasets, stronger evaluation protocols, and
more advanced architectures while preserving the same family-level organization
and consistent training interface.

\section{Conclusion}

We presented \textsc{Venom}, a PyTorch toolkit that brings major generative
modeling paradigms into a single educational and benchmark-friendly codebase.
By implementing representative diffusion, score-based, flow matching, VAE,
normalizing flow, GAN, and EBM methods with consistent training and sampling
interfaces, \textsc{Venom} helps beginners understand the landscape of modern
generative modeling and provides researchers with a compact scaffold for
experimentation, extension, and lightweight comparison.

\section*{Software Availability}

The project page is available at
\url{https://divinyan.com/Venom/}. The source code is available at
\url{https://github.com/yanliang3612/Venom}.

\section*{Acknowledgments}

This software paper draft describes the open-source \textsc{Venom} project.

\bibliographystyle{plainnat}
\bibliography{references}

@incollection{lecun2006energy,
  title = {A Tutorial on Energy-Based Learning},
  author = {LeCun, Yann and Chopra, Sumit and Hadsell, Raia and Ranzato, Marc'Aurelio and Huang, Fu Jie},
  booktitle = {Predicting Structured Data},
  publisher = {MIT Press},
  year = {2006}
}

@inproceedings{ho2020ddpm,
  title = {Denoising Diffusion Probabilistic Models},
  author = {Ho, Jonathan and Jain, Ajay and Abbeel, Pieter},
  booktitle = {Advances in Neural Information Processing Systems},
  year = {2020}
}

@inproceedings{song2020ddim,
  title = {Denoising Diffusion Implicit Models},
  author = {Song, Jiaming and Meng, Chenlin and Ermon, Stefano},
  booktitle = {International Conference on Learning Representations},
  year = {2021}
}

@inproceedings{nichol2021improved,
  title = {Improved Denoising Diffusion Probabilistic Models},
  author = {Nichol, Alexander Quinn and Dhariwal, Prafulla},
  booktitle = {International Conference on Machine Learning},
  year = {2021}
}

@inproceedings{dhariwal2021diffusion,
  title = {Diffusion Models Beat {GANs} on Image Synthesis},
  author = {Dhariwal, Prafulla and Nichol, Alexander Quinn},
  booktitle = {Advances in Neural Information Processing Systems},
  year = {2021}
}

@inproceedings{ho2021classifierfree,
  title = {Classifier-Free Diffusion Guidance},
  author = {Ho, Jonathan and Salimans, Tim},
  booktitle = {NeurIPS Workshop on Deep Generative Models and Downstream Applications},
  year = {2021}
}

@inproceedings{song2019ncsn,
  title = {Generative Modeling by Estimating Gradients of the Data Distribution},
  author = {Song, Yang and Ermon, Stefano},
  booktitle = {Advances in Neural Information Processing Systems},
  year = {2019}
}

@inproceedings{song2021score,
  title = {Score-Based Generative Modeling through Stochastic Differential Equations},
  author = {Song, Yang and Sohl-Dickstein, Jascha and Kingma, Diederik P. and Kumar, Abhishek and Ermon, Stefano and Poole, Ben},
  booktitle = {International Conference on Learning Representations},
  year = {2021}
}

@inproceedings{karras2022edm,
  title = {Elucidating the Design Space of Diffusion-Based Generative Models},
  author = {Karras, Tero and Aittala, Miika and Aila, Timo and Laine, Samuli},
  booktitle = {Advances in Neural Information Processing Systems},
  year = {2022}
}

@inproceedings{lu2022dpmsolver,
  title = {{DPM-Solver}: A Fast {ODE} Solver for Diffusion Probabilistic Model Sampling in Around 10 Steps},
  author = {Lu, Cheng and Zhou, Yuhao and Bao, Fan and Chen, Jianfei and Li, Chongxuan and Zhu, Jun},
  booktitle = {Advances in Neural Information Processing Systems},
  year = {2022}
}

@inproceedings{xu2022pfgm,
  title = {Poisson Flow Generative Models},
  author = {Xu, Yilun and Liu, Ziming and Tegmark, Max and Jaakkola, Tommi},
  booktitle = {Advances in Neural Information Processing Systems},
  year = {2022}
}

@inproceedings{liu2023rectifiedflow,
  title = {Flow Straight and Fast: Learning to Generate and Transfer Data with Rectified Flow},
  author = {Liu, Xingchao and Gong, Chengyue and Liu, Qiang},
  booktitle = {International Conference on Learning Representations},
  year = {2023}
}

@inproceedings{lipman2023flow,
  title = {Flow Matching for Generative Modeling},
  author = {Lipman, Yaron and Chen, Ricky T. Q. and Ben-Hamu, Heli and Nickel, Maximilian and Le, Matt},
  booktitle = {International Conference on Learning Representations},
  year = {2023}
}

@article{tong2024conditionalflowmatching,
  title = {Improving and Generalizing Flow-Based Generative Models with Minibatch Optimal Transport},
  author = {Tong, Alexander and Fatras, Kilian and Malkin, Nikolay and Huguet, Guillaume and Zhang, Yanlei and Rector-Brooks, Jarrid and Wolf, Guy and Bengio, Yoshua},
  journal = {Transactions on Machine Learning Research},
  year = {2024}
}

@inproceedings{salimans2022progressive,
  title = {Progressive Distillation for Fast Sampling of Diffusion Models},
  author = {Salimans, Tim and Ho, Jonathan},
  booktitle = {International Conference on Learning Representations},
  year = {2022}
}

@inproceedings{song2023consistency,
  title = {Consistency Models},
  author = {Song, Yang and Dhariwal, Prafulla and Chen, Mark and Sutskever, Ilya},
  booktitle = {International Conference on Machine Learning},
  year = {2023}
}

@inproceedings{frans2025shortcut,
  title = {One Step Diffusion via Shortcut Models},
  author = {Frans, Kevin and Hafner, Danijar and Levine, Sergey and Abbeel, Pieter},
  booktitle = {International Conference on Learning Representations},
  year = {2025}
}

@article{geng2025meanflow,
  title = {Mean Flows for One-step Generative Modeling},
  author = {Geng, Zhengyang and Deng, Mingyang and Bai, Xingjian and Kolter, J. Zico and He, Kaiming},
  journal = {arXiv preprint arXiv:2505.13447},
  year = {2025}
}

@inproceedings{kingma2014vae,
  title = {Auto-Encoding Variational Bayes},
  author = {Kingma, Diederik P. and Welling, Max},
  booktitle = {International Conference on Learning Representations},
  year = {2014}
}

@inproceedings{burda2016iwae,
  title = {Importance Weighted Autoencoders},
  author = {Burda, Yuri and Grosse, Roger and Salakhutdinov, Ruslan},
  booktitle = {International Conference on Learning Representations},
  year = {2016}
}

@inproceedings{higgins2017beta,
  title = {{beta-VAE}: Learning Basic Visual Concepts with a Constrained Variational Framework},
  author = {Higgins, Irina and Matthey, Loic and Pal, Arka and Burgess, Christopher P. and Glorot, Xavier and Botvinick, Matthew and Mohamed, Shakir and Lerchner, Alexander},
  booktitle = {International Conference on Learning Representations},
  year = {2017}
}

@inproceedings{oord2017vqvae,
  title = {Neural Discrete Representation Learning},
  author = {van den Oord, Aaron and Vinyals, Oriol and Kavukcuoglu, Koray},
  booktitle = {Advances in Neural Information Processing Systems},
  year = {2017}
}

@inproceedings{rezende2015normalizing,
  title = {Variational Inference with Normalizing Flows},
  author = {Rezende, Danilo Jimenez and Mohamed, Shakir},
  booktitle = {International Conference on Machine Learning},
  year = {2015}
}

@article{dinh2014nice,
  title = {{NICE}: Non-linear Independent Components Estimation},
  author = {Dinh, Laurent and Krueger, David and Bengio, Yoshua},
  journal = {arXiv preprint arXiv:1410.8516},
  year = {2014}
}

@inproceedings{dinh2017realnvp,
  title = {Density Estimation using Real {NVP}},
  author = {Dinh, Laurent and Sohl-Dickstein, Jascha and Bengio, Samy},
  booktitle = {International Conference on Learning Representations},
  year = {2017}
}

@inproceedings{kingma2018glow,
  title = {Glow: Generative Flow with Invertible 1x1 Convolutions},
  author = {Kingma, Diederik P. and Dhariwal, Prafulla},
  booktitle = {Advances in Neural Information Processing Systems},
  year = {2018}
}

@inproceedings{papamakarios2017maf,
  title = {Masked Autoregressive Flow for Density Estimation},
  author = {Papamakarios, George and Pavlakou, Theo and Murray, Iain},
  booktitle = {Advances in Neural Information Processing Systems},
  year = {2017}
}

@inproceedings{kingma2016iaf,
  title = {Improved Variational Inference with Inverse Autoregressive Flow},
  author = {Kingma, Diederik P. and Salimans, Tim and Jozefowicz, Rafal and Chen, Xi and Sutskever, Ilya and Welling, Max},
  booktitle = {Advances in Neural Information Processing Systems},
  year = {2016}
}

@inproceedings{durkan2019spline,
  title = {Neural Spline Flows},
  author = {Durkan, Conor and Bekasov, Artur and Murray, Iain and Papamakarios, George},
  booktitle = {Advances in Neural Information Processing Systems},
  year = {2019}
}

@inproceedings{grathwohl2019ffjord,
  title = {{FFJORD}: Free-Form Continuous Dynamics for Scalable Reversible Generative Models},
  author = {Grathwohl, Will and Chen, Ricky T. Q. and Bettencourt, Jesse and Sutskever, Ilya and Duvenaud, David},
  booktitle = {International Conference on Learning Representations},
  year = {2019}
}

@inproceedings{ho2019flowpp,
  title = {Flow++: Improving Flow-Based Generative Models with Variational Dequantization and Architecture Design},
  author = {Ho, Jonathan and Chen, Xi and Srinivas, Aravind and Duan, Yan and Abbeel, Pieter},
  booktitle = {International Conference on Machine Learning},
  year = {2019}
}

@inproceedings{goodfellow2014gan,
  title = {Generative Adversarial Nets},
  author = {Goodfellow, Ian and Pouget-Abadie, Jean and Mirza, Mehdi and Xu, Bing and Warde-Farley, David and Ozair, Sherjil and Courville, Aaron and Bengio, Yoshua},
  booktitle = {Advances in Neural Information Processing Systems},
  year = {2014}
}

@article{mirza2014cgan,
  title = {Conditional Generative Adversarial Nets},
  author = {Mirza, Mehdi and Osindero, Simon},
  journal = {arXiv preprint arXiv:1411.1784},
  year = {2014}
}

@inproceedings{radford2016dcgan,
  title = {Unsupervised Representation Learning with Deep Convolutional Generative Adversarial Networks},
  author = {Radford, Alec and Metz, Luke and Chintala, Soumith},
  booktitle = {International Conference on Learning Representations},
  year = {2016}
}

@inproceedings{odena2017acgan,
  title = {Conditional Image Synthesis with Auxiliary Classifier GANs},
  author = {Odena, Augustus and Olah, Christopher and Shlens, Jonathon},
  booktitle = {International Conference on Machine Learning},
  year = {2017}
}

@inproceedings{chen2016infogan,
  title = {{InfoGAN}: Interpretable Representation Learning by Information Maximizing Generative Adversarial Nets},
  author = {Chen, Xi and Duan, Yan and Houthooft, Rein and Schulman, John and Sutskever, Ilya and Abbeel, Pieter},
  booktitle = {Advances in Neural Information Processing Systems},
  year = {2016}
}

@inproceedings{mao2017lsgan,
  title = {Least Squares Generative Adversarial Networks},
  author = {Mao, Xudong and Li, Qing and Xie, Haoran and Lau, Raymond Y. K. and Wang, Zhen and Smolley, Stephen Paul},
  booktitle = {International Conference on Computer Vision},
  year = {2017}
}

@inproceedings{arjovsky2017wgan,
  title = {Wasserstein Generative Adversarial Networks},
  author = {Arjovsky, Martin and Chintala, Soumith and Bottou, Leon},
  booktitle = {International Conference on Machine Learning},
  year = {2017}
}

@inproceedings{gulrajani2017wgangp,
  title = {Improved Training of Wasserstein {GANs}},
  author = {Gulrajani, Ishaan and Ahmed, Faruk and Arjovsky, Martin and Dumoulin, Vincent and Courville, Aaron},
  booktitle = {Advances in Neural Information Processing Systems},
  year = {2017}
}

@article{lim2017geometricgan,
  title = {Geometric {GAN}},
  author = {Lim, Jae Hyun and Ye, Jong Chul},
  journal = {arXiv preprint arXiv:1705.02894},
  year = {2017}
}

@inproceedings{miyato2018sngan,
  title = {Spectral Normalization for Generative Adversarial Networks},
  author = {Miyato, Takeru and Kataoka, Toshiki and Koyama, Masanori and Yoshida, Yuichi},
  booktitle = {International Conference on Learning Representations},
  year = {2018}
}

@article{hinton2002cd,
  title = {Training Products of Experts by Minimizing Contrastive Divergence},
  author = {Hinton, Geoffrey E.},
  journal = {Neural Computation},
  volume = {14},
  number = {8},
  pages = {1771--1800},
  year = {2002}
}

@article{hyvarinen2005score,
  title = {Estimation of Non-Normalized Statistical Models by Score Matching},
  author = {Hyv{\"a}rinen, Aapo},
  journal = {Journal of Machine Learning Research},
  volume = {6},
  pages = {695--709},
  year = {2005}
}

@inproceedings{gutmann2010nce,
  title = {Noise-Contrastive Estimation: A New Estimation Principle for Unnormalized Statistical Models},
  author = {Gutmann, Michael and Hyv{\"a}rinen, Aapo},
  booktitle = {International Conference on Artificial Intelligence and Statistics},
  year = {2010}
}

@inproceedings{du2019implicit,
  title = {Implicit Generation and Modeling with Energy Based Models},
  author = {Du, Yilun and Mordatch, Igor},
  booktitle = {Advances in Neural Information Processing Systems},
  year = {2019}
}

@inproceedings{grathwohl2020jem,
  title = {Your Classifier is Secretly an Energy Based Model and You Should Treat it Like One},
  author = {Grathwohl, Will and Wang, Kuan-Chieh and Jacobsen, J{\"o}rn-Henrik and Duvenaud, David and Norouzi, Mohammad and Swersky, Kevin},
  booktitle = {International Conference on Learning Representations},
  year = {2020}
}

@article{albergo2023stochastic,
  title = {Stochastic Interpolants: A Unifying Framework for Flows and Diffusions},
  author = {Albergo, Michael S. and Boffi, Nicholas M. and Vanden-Eijnden, Eric},
  journal = {arXiv preprint arXiv:2303.08797},
  year = {2023}
}

\end{document}